\documentclass{esannV2}
\usepackage[]{hyperref}
\usepackage[dvips]{graphicx}
\usepackage[utf8]{inputenc}
\usepackage{amssymb,amsmath,array,booktabs,dsfont}
\DeclareMathOperator*{\argmax}{arg\,max}

\begin{document}
\title{Improving the Linearized Laplace Approximation via Quadratic Approximations}


\author{Pedro Jiménez$^1$, Luis A. Ortega$^{1,2}$, \\Pablo Morales-Álvarez$^{3,4}$, Daniel Hernández-Lobato$^{1,5}$
%
\thanks{DHL and LAO acknowledge financial support from project PID2022-139856NB-I00 funded by MCIN, and from project IDEA-CM (TEC-2024/COM-89) and the ELLIS Unit Madrid, funded
by the Autonomous Community of Madrid. We acknowledge support from Centro de Computaci\'on Cient\'ifica-Universidad Aut\'onoma de Madrid (CCC-UAM). PMA acknowledges project PID2022-140189OB-C22 funded by MCIN/AEI/10.13039/501100011033 (Spanish Ministry of Science) and grant C-EXP-153-UGR23 funded by Consejer\'ia de Universidad, Investigaci\'on e Innovaci\'on and by the European Union (EU) ERDF Andalusia Program 2021–2027.}
%
\vspace{.3cm}\\
%
1- Machine Learning Group - Computer Science Department \\
Escuela Polit\'ecnica Superior, Universidad Aut\'onoma de Madrid, Spain
%
\vspace{.1cm}\\
2- Department of Computer Science, Aalborg University, Copenhagen
\vspace{.1cm}\\
3- Visual Information Processing Group \\
Department of Statistics and Operation Research, University of Granada, Spain
\vspace{.1cm}\\
4- Information and Communication Technologies Research Centre (CITIC),\\
University of Granada, Spain
\vspace{.1cm}\\
5- Centro de Investigaci\'on Avanzada en F\'isica Fundamental, \\Universidad Aut\'onoma de Madrid, Spain
}


\maketitle

\begin{abstract}
Deep neural networks (DNNs) often produce overconfident out-of-distribution predictions, 
motivating Bayesian uncertainty quantification. The Linearized Laplace Approximation (LLA) 
achieves this by linearizing the DNN and applying Laplace inference to the resulting 
model. Importantly, the linear model is also used for prediction. 
We argue this linearization in the posterior may degrade
fidelity to the true Laplace approximation. To alleviate this problem, without increasing 
significantly the computational cost, we propose the Quadratic Laplace Approximation (QLA). 
QLA approximates each second order factor in the approximate Laplace log-posterior using 
a rank-one factor obtained via efficient power iterations. QLA is expected to yield 
a posterior precision closer to that of the full Laplace without forming the full 
Hessian, which is typically intractable. For prediction, QLA also uses the linearized model.
Empirically, QLA yields modest yet 
consistent uncertainty estimation improvements over LLA on five regression datasets.
\end{abstract}

\section{Introduction}

Deep neural networks (DNNs) are the state-of-the-art models for many supervised learning tasks, 
achieving remarkable performance across different domains \cite{LeCun2015}. Despite their predictive 
power, such models often produce over-confident outputs for inputs that lie far from the training 
distribution, which undermines their reliability in safety-critical applications 
\cite{ortega2024variationallinearizedlaplaceapproximation}. A principled remedy is to adopt a 
Bayesian treatment of the model parameters, which yields a posterior distribution over the parameters 
and, via marginalization, a predictive distribution that quantifies uncertainty alongside point predictions \cite{Bishop}.

Exact Bayesian inference is intractable for modern DNNs due to the high-dimensional integrals 
involved and the large number of parameters. Consequently, approximate inference methods are 
widely used. The Laplace approximation (LA) \cite{MacKay_BayesModelComp} is classical and 
conceptually simple but becomes computationally demanding in deep models because it requires 
handling very large Hessians. The Linearized Laplace Approximation (LLA) \cite{immer21a} 
linearizes the network around the parameters and applies LA in the resulting linear 
model. LLA substantially reduces cost and yields useful predictive distributions in 
practice. Importantly, using the linear model for prediction also alleviates some of the 
biases of LA \cite{immer21a}.

Linearization in the posterior computation, however, may reduce expressivity and degrade 
uncertainty calibration. To better balance efficiency and fidelity, we propose the \emph{Quadratic Laplace Approximation} (QLA). 
QLA builds a second-order expansion of the network around the maximum a posteriori estimation (MAP). However, it approximates 
the dominant per-datapoint precision terms by a rank-one factor obtained via power iterations with 
efficient Hessian-vector products. This yields \emph{refined Jacobians} without forming the full Hessian. 
For prediction, we employ the linearized predictive model, which, as shown in \cite{immer21a}, helps to mitigate the 
effect of the tendency to assign excessive probability mass to regions of parameter space unsupported by data. 
Empirically, QLA produces small but consistent improvements in uncertainty estimation over LLA across five regression datasets.

\section{Linearized Laplace Approximation}

We consider a regression problem with data $\mathcal{D}=\{(\mathbf{x}_1,y_1),\dotsc,(\mathbf{x}_N,y_N)\}$, 
where each input $\mathbf{x}_n\in\mathbb{R}^D$ and target $y_n\in\mathbb{R}$. A neural networks is a 
parametric function $f:\mathbb{R}^D\times\mathbb{R}^P\to\mathbb{R}$ with $P$ learnable parameters, 
mapping an input $\mathbf{x}_n$ and a parameter vector $\theta\in\mathbb{R}^P$ to 
a predicted output $f\left(\mathbf{x}_n,\theta\right)$.

The target variable is modeled as the network output plus random noise, inducing the conditional 
observation model $p\left(y\mid f\left(\mathbf{x},\theta\right)\right)$. Assuming conditional 
independence across samples, the likelihood factorizes as
\begin{align}
	p\left(\mathcal{D}\mid\theta\right) & = \textstyle \prod_{n=1}^N p\left(y_n\mid f\left(\mathbf{x}_n,\theta\right)\right)\,.
\end{align}
Given a prior distribution $p(\theta)$, a Bayesian treatment aims to infer the posterior over 
parameters conditioned on the observed data. However, exact inference is intractable in modern 
DNNs due to the high dimensionality of $\theta$ and the non-linearity of $f$. We therefore resort 
to LA, which approximates the posterior by a Gaussian $q(\theta)$ centered at the MAP estimate $\theta^\ast$:
\begin{align}
	q(\theta) & =\textstyle \mathcal{N}\left(\theta\mid\theta^\ast,\Sigma\right),\qquad \theta^\ast=\argmax_{\theta}\ell\left(\theta,\mathcal{D}\right)\,,
\end{align}
where $\ell\left(\theta,\mathcal{D}\right)=\sum_{n=1}^N\log p\left(y_n\mid f\left(\mathbf{x}_n,\theta\right)\right)+\log p(\theta)$ 
is the log-joint density. The precision matrix is given by the negative Hessian at $\theta^\ast$:
$\Sigma^{-1}=-\nabla_{\theta}^2 \ell\left(\theta,\mathcal{D}\right)\Big|_{\theta=\theta^\ast}$.
To compute this Hessian we use its per-datum decomposition,
\begin{align}
	\nabla_{\theta}^2 \ell\left(\theta,\mathcal{D}\right) & =\textstyle \sum_{n=1}^N\nabla_{\theta}^2\log p\left(y_n\mid f\left(\mathbf{x}_n,\theta\right)\right)+\nabla_{\theta}^2 \log p(\theta)\,.
\end{align}
The prior is typically tractable, so we focus on the data-dependent summands:
\begin{align}
\nabla_{\theta}\log p\left(y\mid f\left(\mathbf{x},\theta\right)\right)&=\mathbf{r}\left(y;f\right)\mathcal{J}_{\theta}(\mathbf{x})\,,\\
\nabla_{\theta}^2\log p\left(y\mid f\left(\mathbf{x},\theta\right)\right)&=\mathbf{r}\left(y;f\right)\mathcal{H}_{\theta}(\mathbf{x})-\mathcal{J}_{\theta}(\mathbf{x}) \Lambda(y;f)\mathcal{J}_{\theta}^\intercal(\mathbf{x})\,,
\end{align}
where $\mathcal{J}_{\theta}(\mathbf{x})$ and $\mathcal{H}_{\theta}(\mathbf{x})$ are the parameter Jacobian 
and parameter Hessian of the network, $\mathbf{r}\left(y;f\right)$ denotes the residual, and $ \Lambda(y;f)$ denotes 
per-input noise. The generalized Gauss-Newton (GGN) approximation \cite{pmlr-v37-martens15} discards the term 
$\mathcal{H}_{\theta}(\mathbf{x})^\intercal\mathbf{r}\left(y;f\right)$, yielding
\begin{align}
	\nabla_{\theta}^2\log p\left(y\mid f\left(\mathbf{x},\theta\right)\right) &\approx-\mathcal{J}_{\theta}(\mathbf{x})\Lambda(y;f)\mathcal{J}_{\theta}^\intercal(\mathbf{x})\,.
\end{align}
This approximation can be seen as a linearization of the DNN around $\theta^\ast$ \cite{immer21a},
\begin{align}
	f\left(\mathbf{x},\theta\right) &\approx f_{\text{lin}}^{\theta^\ast} (\mathbf{x},\theta)= f (\mathbf{x},\theta^\ast) + \mathcal{J}_{\theta^\ast}^\intercal(\mathbf{x})(\theta-\theta^\ast)\,.
\end{align}
Applying LA to the linearized model, the posterior approximation becomes
\begin{align}
	q(\theta)& =\mathcal{N}\left(\theta\mid\theta^\ast,\Sigma_{\textbf{GGN}}\right)\,, & \Sigma_{\textbf{GGN}}^{-1} & =\textstyle \sum_{n=1}^N \mathcal{J}_{\theta^\ast}(\mathbf{x}_n) \Lambda(y_n;f_n)\mathcal{J}_{\theta^\ast}^\intercal(\mathbf{x}_n)+\mathbf{S}_0^{-1}\,, \nonumber
\end{align}
with $\mathbf{S}_0$ the prior covariance, typically diagonal. Marginalizing under $q(\theta)$ yields the standard Laplace predictive, 
\begin{align}
	p\left(y\mid\mathbf{x},\mathcal{D}\right)&=\mathds{E}_{q(\theta)}\left[p\left(y\mid f\left(\mathbf{x},\theta\right)\right)\right]\,.
\end{align}
Following \cite{immer21a}, one may instead use the linearized model for prediction to get better results, giving the generalized linear model (GLM) predictive
\begin{align}
	p_{\text{GLM}}\left(y\mid\mathbf{x},\mathcal{D}\right) & = \textstyle \mathds{E}_{q(\theta)}\left[p\left(y\mid f_{\text{lin}}^{\theta^\ast}\left(\mathbf{x},\theta\right)\right)\right]\,.
\end{align}
For Gaussian observations, this reduces to a Gaussian predictive distribution 
with mean $f\left(\mathbf{x},\theta^\ast\right)$ and variance $\mathcal{J}_{\theta^\ast}^\intercal(\mathbf{x})\Sigma_{\textbf{GGN}} \mathcal{J}_{\theta^\ast}(\mathbf{x})$.

\section{Quadratic Laplace Approximation}

LLA provides an efficient framework that substantially simplifies the original LA computation. By locally approximating the DNN
with a linear model, LLA avoids the explicit computation of Hessians and, when using the linearized model for predictive 
inference, yields more stable uncertainty estimates \cite{immer21a}.

The linearity assumption in the posterior, however, may bias the posterior variance estimation, which can 
limit the fidelity of prediction uncertainty estimation. To address this limitation, we propose the 
\emph{Quadratic Laplace Approximation} (QLA), which extends LLA by adopting a quadratic local approximation of the DNN
in the posterior computation:
\begin{align}
f\left(\mathbf{x},\theta\right)\approx f_{\text{quad}}^{\theta^\ast}\left(\mathbf{x},\theta\right)=
	f_{\text{lin}}^{\theta^\ast}\left(\mathbf{x},\theta\right) +0.5\left(\theta-\theta^\ast\right)^\intercal\mathcal{H}_{\theta^\ast}(\mathbf{x})\left(\theta-\theta^\ast\right)\,,
\end{align}
where $\mathcal{J}_{\theta^\ast}(\mathbf{x})$ and $\mathcal{H}_{\theta^\ast}(\mathbf{x})$ denote the Jacobian and Hessian of the network evaluated at the MAP estimate $\theta^\ast$. This corresponds to a \emph{Quadratic Taylor Expansion} (QTE) around $\theta^\ast$. This is the approach followed in LA.

As in the classical LA, we approximate the posterior by a Gaussian centered at the MAP solution:
\begin{align}
	q(\theta)=& \mathcal{N}\left(\theta\mid\theta^\ast,\Sigma_{\textbf{QTE}}\right)\,, 
\end{align}
with precision
\begin{align}
	\Sigma_{\textbf{QTE}}^{-1}& =\textstyle \sum_{n=1}^N \nabla_{\theta}^2\log p\left(y_n\mid f_{\text{quad}}^{\theta^\ast}\left(\mathbf{x}_n,\theta\right)\right)+\mathbf{S}_0^{-1}\,.
\end{align}
Focusing on the data-dependent term, we obtain
\begin{align}
\textstyle 
\nabla_{\theta}^2\log p\!\left(y\mid f_{\text{quad}}^{\theta^\ast}\left(\mathbf{x},\theta\right)\right)\Big|_{\theta=\theta^\ast}=\mathbf{r}(y;f_{\text{quad}}^{\theta^\ast})\mathcal{H}_{\theta^\ast}(\mathbf{x})-\mathcal{J}_{\theta^\ast}(\mathbf{x}) \Lambda(y;f_{\text{quad}}^{\theta^\ast})\mathcal{J}_{\theta^\ast}^\intercal(\mathbf{x})\,.
\nonumber
\end{align}
At this stage, we face the challenge of explicitly obtaining the network Hessian. However, in frameworks such as \emph{PyTorch}, Hessian-vector products can be efficiently computed without forming the full Hessian matrix. Exploiting this property, we employ the \emph{power iteration} method to estimate the dominant eigenvector of
\begin{align}
	\mathbf{A} & :=\textstyle \nabla_{\theta}^2\log p\!\left(y\mid f_{\text{quad}}^{\theta^\ast}\left(\mathbf{x},\theta\right)\right)\Big|_{\theta=\theta^\ast}\,. 
\end{align}
We initialize the procedure with the network Jacobian at the reference point, $\mathbf{z}^{(0)}=\mathcal{J}_{\theta^\ast}(\mathbf{x})$, and recursively update
\begin{align}
	\mathbf{z}^{(k+1)} & =\frac{\mathbf{A}\mathbf{z}^{(k)}}{||\mathbf{A}\mathbf{z}^{(k)}||}\,.
\end{align}
By distributivity, in each multiplication by $\mathbf{A}$, $\mathbf{A}\mathbf{z}^{(k)}$ decomposes as
\begin{align}
\mathbf{A}\mathbf{z}^{(k)} & =\mathbf{r}(y;f_{\text{quad}}^{\theta^\ast})\mathcal{H}_{\theta^\ast}(\mathbf{x})\mathbf{z}^{(k)}-\mathcal{J}_{\theta^\ast}(\mathbf{x}) \Lambda(y;f_{\text{quad}}^{\theta^\ast})\mathcal{J}_{\theta^\ast}^\intercal(\mathbf{x})\mathbf{z}^{(k)}\,.
\end{align}
Thus, the full Hessian never needs to be constructed, only Hessian-vector products 
$\mathcal{H}_{\theta^\ast}(\mathbf{x})\mathbf{z}^{(k)}$ need to be computed, which can be 
efficiently obtained using automatic differentiation.

After a few iterations, 
the algorithm converges to the dominant eigenvector, denoted $\hat{\mathbf{z}}$, which can be 
interpreted as a \emph{refined Jacobian} capturing both first-order information and the 
most significant quadratic contribution. Under this perspective, we approximate 
$\mathbf{A}$ by a rank-one factorization $\mathbf{A}\approx\hat{\mathbf{z}}\hat{\mathbf{z}}^\intercal$, 
leading to the following expression for the Gaussian precision matrix:
\begin{align}
	\Sigma_{\textbf{QTE}}^{-1}& \approx \textstyle \sum_{n=1}^N \hat{\mathbf{z}}_n\hat{\mathbf{z}}_n^\intercal+\mathbf{S}_0^{-1}\,.
\end{align}
Marginalizing under $q(\theta)$ and using the quadratic network approximation yields the approximate predictive distribution
\begin{align}
	p_{\text{QTE}}\left(y\mid\mathbf{x},\mathcal{D}\right) & =\textstyle \mathds{E}_{q(\theta)}\left[p\left(y\mid f_{\text{quad}}^{\theta^\ast}\left(\mathbf{x},\theta\right)\right)\right]\,.
\end{align}
For Gaussian observations, the predictive mean is 
\begin{align}
f (\mathbf{x},\theta^*)+\frac{1}{2}\text{tr}\left(\Sigma_{\textbf{QTE}}\mathcal{H}_{\theta^*}(\mathbf{x})\right)\,,
\end{align}
and the predictive variance is 
\begin{align}
\mathcal{J}_{\theta^*}^\intercal(\mathbf{x})\Sigma_{\textbf{QTE}} 
	\mathcal{J}_{\theta^*}(\mathbf{x})+\frac{1}{2}\text{tr}\left[\left({\mathcal{H}}_{\theta^*}(\mathbf{x}_i)\Sigma_{\textbf{QTE}}\right)^2\right]\,.
\end{align}
Two drawbacks arise at this stage. First, the predictive mean no longer coincides with the network 
output evaluated at the MAP parameters. Second, the resulting variance, being closer in form to 
that of the standard LA, does not sufficiently mitigate the approximation's tendency to assign 
excessive probability mass to unsupported parameter regions.

To address this, we instead use the linearized model for predictions, as in LLA, which represents the best 
first-order approximation to our quadratic expansion. The resulting predictive distribution is then given by
\begin{align}
	p_{\text{QTE}}\left(y\mid\mathbf{x},\mathcal{D}\right)& =\mathcal{N}\left(y\mid f\left(\mathbf{x},\theta^\ast\right),\mathcal{J}_{\theta^\ast}^\intercal(\mathbf{x})\Sigma_{\textbf{QTE}} \mathcal{J}_{\theta^\ast}(\mathbf{x})\right)\,.
\end{align}

\vspace{-4mm}
\section{Experiments}
We compare LLA and QLA on standard UCI regression datasets: Boston Housing, Energy Efficiency, Yacht Hydrodynamics, 
Concrete Compressive Strength and Wine Quality. Inputs and targets were standardized prior to training. 
To assess uncertainty estimation in out-of-distribution samples, we use the \emph{in-between} splits proposed in
\cite{foong2019inbetweenuncertaintybayesianneural}, see also \cite{morales-alvarez2021activationlevel}. For each input dimension, the data are sorted, 
the middle third is held out as test set and the outer two thirds form the training set, yielding $D$ gap splits 
per dataset.

For each dataset and each gap split we train one DNN using back-propagation (thus $D$ networks per dataset). 
The DNN hyper-parameters (weight decay $\{0.0, 10^{-4}, 10^{-3}$\}, number of units $\{20, 30, 50\}$, and number of layers $\{1,2,3\}$) are found using an inner-cv method on the training set. Then, we obtain the predictive distributions using both LLA and QLA, and compute the chosen metrics on the corresponding test split.  In QLA, we use the prior variance estimated by LLA. The estimation is performed by maximizing the marginal likelihood as in \cite{immer21a}. 
We use $10$ iterations of the power method in QLA. Finally, we report metrics averaged across data splits.

The quality of the predictive distribution is measured using the Negative Log Likelihood (NLL), and the Continuous Ranked 
Probability Score (CRPS) \cite{Gneiting01032007}. These are popular metrics used in the literature involving LLA and uncertainty estimation \cite{ortega2024variationallinearizedlaplaceapproximation}. Because LLA and QLA share the same predictive mean, differences in these metrics reflect differences in predictive variance and therefore directly 
assess each method's ability to produce realistic uncertainty estimates.

Full obtained results are reported in Table~\ref{tab:results_compact}. The table shows average test scores (NLL and CRPS, the lower the better) for LLA and QLA, aggregated over the in-between splits, for each dataset. Best results are highlighted in bold. Overall, QLA yields comparable performance to LLA and produces small, but consistent improvements in the uncertainty metrics across most datasets, although the differences are generally small. These results indicate that QLA refines the posterior precision and modestly improves uncertainty estimation while conserving the predictive performance of the initial DNN.
\vspace{-4mm}

\begin{table}[htbp]
\centering
\small
\caption{Average metric results across each dataset for LLA and QLA.}\label{tab:results_compact}
    \centering
    \small
	\begin{tabular*}{\linewidth}{@{\extracolsep{\fill}} l@{\hspace{1mm}} c@{\hspace{1mm}} c@{\hspace{1mm}}  c@{\hspace{1mm}} c@{\hspace{1mm}}c@{\hspace{1mm}}c}
\toprule
Dataset & \multicolumn{2}{c}{NLL ($\downarrow$)} & \multicolumn{2}{c}{CRPS ($\downarrow$)} \\
\cmidrule(lr){2-3}\cmidrule(lr){4-5}\cmidrule(lr){6-7}
 & LLA & QLA & LLA & QLA \\
\midrule
Boston   & $2.7290\pm 0.0402$ & $\bf 2.7279\pm 0.0401$
        & $2.0581\pm 0.0941$ & $\bf 2.0579\pm 0.0941$ \\
Energy   & $2.8704\pm 0.4524$ & $\bf 2.8387\pm 0.4455$
         & $2.5091\pm 0.8006$ & $\bf 2.5009\pm 0.7975$ \\
Yacht    & $2.5153\pm 0.0899$ & $\bf 2.5143\pm 0.0887$
         & $\bf 1.6166\pm 0.2008$ & $1.6169\pm 0.1999$ \\
Concrete & $3.3332\pm 0.0227$ & $\bf 3.3331\pm 0.0228$
         & $3.8750\pm 0.1003$ & $\bf 3.8742\pm 0.1003$ \\
Wine     & $1.0022\pm 0.0089$ & $\bf 1.0015\pm 0.0089$
         & $0.3663\pm 0.0040$ & $\bf 0.3661\pm 0.0040$ \\
\bottomrule
\end{tabular*}
\end{table}


\vspace{-5mm}
\section{Conclusions}
\vspace{-2mm}
In this paper we have introduced the \emph{Quadratic Laplace Approximation} (QLA), an extension of the 
Linearized Laplace Approximation that incorporates second-order terms in the log-posterior through 
efficient rank-one updates obtained via the power iteration method. This yields a refined posterior 
precision without explicit Hessian computation. For prediction, it retains stable predictive behavior of LLA 
through the linearized predictive model, which mitigates the tendency to allocate excessive probability mass to 
unsupported parameter regions \cite{immer21a}. Experiments on five regression datasets show that QLA provides small 
but consistent improvements in uncertainty metrics over LLA at negligible additional cost. Future work 
will address extending QLA to classification and multivariate regression, and improving scalability through 
low-rank or inducing-point based strategies for large neural networks \cite{ortega2024variationallinearizedlaplaceapproximation}.


\begin{footnotesize}

\bibliographystyle{unsrt}
\bibliography{library.bib}

@book{Bishop, place={New York}, title={Pattern Recognition and Machine Learning}, publisher={Springer New York, NY}, author={Bishop, C. M.}, year={2006}}

@InProceedings{immer21a,
  title = 	 { Improving predictions of {B}ayesian neural nets via local linearization },
  author =       {Immer, A. and Korzepa, M. and Bauer, M.},
  booktitle = 	 {AISTATS},
  year = 	 {2021},
}

@inproceedings{MacKay_BayesModelComp,
author = {MacKay, D. J. C.},
title = {Bayesian model comparison and backprop nets},
year = {1991},
isbn = {1558602224},
booktitle = {Proceedings of the 5th International Conference on Neural Information Processing Systems},
numpages = {8},
location = {Denver, Colorado},
}

@inproceedings{ortega2024variationallinearizedlaplaceapproximation,
  author       = {L. A. Ortega  and
                  S. Rodr{\'{\i}}guez-Santana and
                  D. Hern{\'{a}}ndez{-}Lobato},
  title        = {Variational Linearized Laplace Approximation for {B}ayesian Deep Learning},
  booktitle    = {ICML},
  year         = {2024},
}

@article{Gneiting01032007,
author = {T. Gneiting and A. E. Raftery},
title = {Strictly Proper Scoring Rules, Prediction, and Estimation},
journal = {Journal of the American Statistical Association},
volume = {102},
pages = {359--378},
year = {2007},
}

@InProceedings{pmlr-v37-martens15,
  title = 	 {Optimizing Neural Networks with Kronecker-factored Approximate Curvature},
  author = 	 {Martens, J. and Grosse, R.},
  booktitle = 	 {ICML},
  year = 	 {2015},
}

@Article{LeCun2015,
author={LeCun, Y.
and Bengio, Y.
and Hinton, G. },
title={Deep learning},
journal={Nature},
year={2015},
volume={521},
pages={436-444},
issn={1476-4687},
doi={10.1038/nature14539},
url={https://doi.org/10.1038/nature14539}
}

@article{foong2019inbetweenuncertaintybayesianneural,
  author       = {A. Y. K. Foong and
                  Y. Li and
                  J. M. Hern{\'{a}}ndez{-}Lobato and
                  R.E. Turner},
  title        = {'{I}n-Between' Uncertainty in Bayesian Neural Networks},
  journal      = {Uncertainty and Robustness in Deep Learning},
  year         = {2019},
}

@inproceedings{morales-alvarez2021activationlevel,
title={Activation-level uncertainty in deep neural networks},
author={P. Morales-Alvarez and D. Hern{\'a}ndez-Lobato and R. Molina and J. M. Hern{\'a}ndez-Lobato},
booktitle={ICLR},
year={2021},
url={https://openreview.net/forum?id=UvBPbpvHRj-}
}

\end{footnotesize}


\end{document}